\documentclass[10pt, journal]{IEEEtran}
\usepackage{pdfpages, amsmath, mathrsfs}

\usepackage{subfig}
\newcommand\circlednumber[1]{\raisebox{0.5pt}{\textcircled{\raisebox{-.9pt} {#1}}}}

\begin{document}

\title{DeepFix: A Fully Convolutional Neural Network for predicting Human Eye Fixations}

\author{Srinivas S S Kruthiventi,
        Kumar Ayush,
        and~R.~Venkatesh~Babu,~\IEEEmembership{Senior~Member,~IEEE}

\thanks{Srinivas S S Kruthiventi and R. Venkatesh Babu are with the Video Analytics Lab, Supercomputer Education
and Research Centre, Indian Institute of Science, Bangalore 560012, India (e-mail: kssaisrinivas@gmail.com; venky@serc.iisc.ernet.in).

Kumar Ayush is with the Indian Institute of Technology Kharagpur, India (email: kumar.ayush@iitkgp.ac.in).}}

\maketitle

\begin{abstract}

Understanding and predicting the human visual attentional mechanism is an active area of research in the fields of neuroscience and computer vision. In this work, we propose DeepFix, a first-of-its-kind fully convolutional neural network for accurate saliency prediction. Unlike classical works which characterize the saliency map using various hand-crafted features, our model automatically learns features in a hierarchical fashion and predicts saliency map in an end-to-end manner. DeepFix is designed to capture semantics at multiple scales while taking global context into account using network layers with very large receptive fields. Generally, fully convolutional nets are spatially invariant which prevents them from modeling location dependent patterns (e.g. centre-bias). Our network overcomes this limitation by incorporating a novel Location Biased Convolutional layer. We evaluate our model on two challenging eye fixation datasets -- MIT300, CAT2000 and show that it outperforms other recent approaches by a 
significant margin.\end{abstract}

\IEEEpeerreviewmaketitle

\section{Introduction}
\label{sec:intro}

Identifying conspicuous stimuli in the visual field is a key attentional mechanism in humans. While free viewing, our eyes tend to fixate on regions of the scene which have distinctive variations in visual stimuli such as a bright colour, unique texture or more complex semantic aspects such as presence of a familiar face or any sudden movements. This mechanism guides our eye gaze to the salient and informative regions in the scene. 

The human visual system is dictated by two kinds of attentional mechanisms: bottom-up and top-down~\cite{connor2004visual}. Bottom-up factors, which are derived entirely from the visual scene, are responsible for the automatic deployment of attention towards discriminative regions in the scene. The involuntary detection of a red coloured \textit{STOP} sign on the road, while driving, is an example of this attentional mechanism. This kind of attention is automatic, reflexive and stimulus-driven. On the contrary, the top-down attention mechanism is driven by internal factors such as subject's prior knowledge, expectations and the task at hand, making it situational and highly subjective~\cite{frintrop2010computational}. It uses information available in the working memory, thereby biasing attention towards areas of the scene important to the current behavioral goals~\cite{awh2006interactions}. The selective attention exhibited by a hungry animal while searching for its camouflaged prey is an example of the top-
down mechanism.

\begin{figure}
\centering
\subfloat{\includegraphics[width = \linewidth]{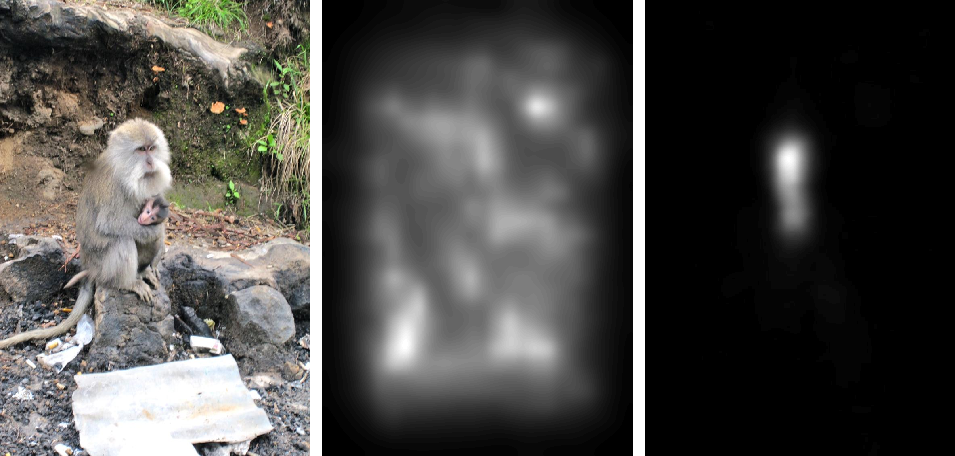}}\\\vspace{-0.1cm}
\subfloat{\includegraphics[width = \linewidth]{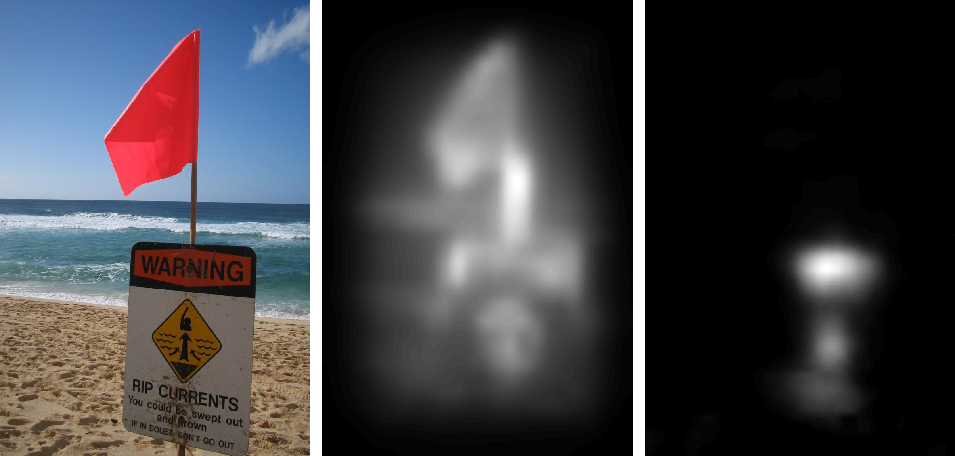}}\\
\caption{Illustrative images (I column) along with saliency maps predicted by a classical method~\cite{harel2006graph} (II column) and DeepFix (III column). For the Monkeys' image in I row, prediction models attribute saliency to the bright white paper and the orange mushrooms in the background leaving out the actual salient region - the monkeys' faces~\cite{Judd_2012}. In the image in II row, even though observers consistently fixate on the text on the board, popular models incorrectly predict saliency in the regions of the bright colored flag and the pole~\cite{Judd_2012}. Our network predicts saliency accurately in both the cases.}
\label{fig:saliencyExample}
\end{figure}

In this work, we propose an approach for modeling the bottom-up visual attentional mechanism by predicting human eye fixations on images. This modeling, commonly referred to as visual saliency prediction, is a classic research area in the field of computer vision and neuroscience~\cite{treisman1980feature, ittiModel}. This modeling has applications in vision-related tasks such as video compression~\cite{hadizadeh2014saliency}, object and action recognition~\cite{walther2002attentional, rapantzikos2009dense}, image retargetting~\cite{chen2003visual} and surveillance systems~\cite{yubing2011spatiotemporal}. In the past, many computational models have been developed to predict human eye fixations in the form of a saliency map -``a topographically arranged two dimensional map that represents the visual saliency of a scene"~\cite{Niebur:2007}. Saliency map predictions for two example images are shown in Fig.~\ref{fig:saliencyExample}.

 Saliency in a visual scene can arise from a spectrum of stimuli, both low-level (color/intensity, orientation, size etc.) and high-level  (faces, text etc.). Most of the classic saliency models~\cite{ittiModel, harel2006graph} are biologically inspired and use multi-scale low-level visual features such as color and texture. However, these methods do not adequately capture the high level semantic aspects of a scene that can contribute to visual saliency. The wide variety of possible causes, both low-level and high-level, make it difficult to hand-craft good features for predicting saliency. This makes deep networks, which are capable of learning features from data in a task dependent manner, a natural choice for this problem. 
 
 Recently, deep networks have shown impressive results on a diverse set of perceptual tasks such as speech recognition~\cite{hinton2012deep}, natural language processing~\cite{collobert2008unified} and object recognition~\cite{krizhevsky2012imagenet}. The ability of deep neural networks to automatically learn complex patterns from data in a hierarchical fashion makes them applicable to a wide range of problems with different modalities of data. Though neural networks are being used in the field of artificial intelligence since many decades, their recent wide applicability can be attributed to the increased computational power of GPUs, efficient techniques for training~\cite{bottou2010large} and the availability of very large datasets~\cite{ILSVRC15, zhou2014learning}.

In this work, we propose a fully convolutional neural network - DeepFix, for predicting human eye fixations on images in the form of a saliency map. Our model, inspired from VGG net~\cite{simonyan2014very}, is a very deep network with 20 convolutional layers, each of a small kernel size, operating in succession on an image. The network is designed to capture object-level semantics, which can occur at multiple scales, efficiently through inception style~\cite{szegedy2014going} convolution blocks. Each inception module consists of a set of convolution layers with different kernel sizes operating in parallel. The global context of the scene, which is crucial for saliency prediction, is captured using convolutional layers with very large receptive fields. These layers are placed towards the end of the network and replace the densely connected inner product layers commonly present in convolutional nets.

Fully Convolutional Nets (FCNs), in general, are location invariant i.e., a spatial shift in the input results only in a corresponding spatial shift of the output without affecting its values. This property prevents FCNs from learning location specific patterns such as the centre-bias. The proposed DeepFix model is designed to handle this through a novel Location Biased Convolutional (LBC) layer.

Overall, our model predicts the saliency map from the image in an end-to-end manner, without relying on any hand-crafted features. Here, we summarize the key aspects of our DeepFix network:

\begin{itemize}
\item Large depth - to enable the extraction of complex semantic features 
\item Kernels of different sizes operating in parallel - to characterize the object semantics simultaneously at multiple scales
\item Kernels with large receptive fields - for capturing the global context 
\item Location biased convolutional layers - for learning location dependent patterns such as the centre-bias present in eye fixations
\end{itemize}

We evaluate the proposed network on two challenging eye fixation datasets -- MIT300~\cite{Judd_2012}, CAT2000~\cite{CAT2000} and show that it achieves state-of-the-art results on both these datasets.

\section{Related Work}
\label{sec.relatedWork}

Treisman \textit{et al.}, in their seminal work of Feature Integration Theory (FIT)~\cite{treisman1980feature}, proposed that preliminary features from visual stimuli are simultaneously processed in different areas of the brain resulting in multiple feature maps. These feature maps are later aggregated to aid in object recognition. Using these principles of human vision, Koch \textit{et al.}~\cite{koch1985shifts} first proposed a biologically plausible computational architecture for modelling these early feature representations to artificially simulate the selective attention mechanism in humans. Later, Itti \textit{et al.}~\cite{ittiModel}, building upon the work of ~\cite{koch1985shifts}, implemented a novel system for visual saliency prediction in images. Their model extracts bottom-up visual features of color, intensity and orientation using various blocks. These features are integrated using a dynamic neural network, to construct a two-dimensional representation, called a saliency map which indicates 
the conspicuity of every pixel in the image. Experimentally, this model was shown to be reasonably successful in detecting centre-surround saliency in images. However, generalization of this model for saliency prediction in complex scenes is difficult because of the primitive nature of its features and the multiple parameters used for constructing them at various scales. Recently, several other works have employed more complex features maps such as isocentric curvedness~\cite{valenti2009computational}, regional histograms~\cite{liu2013saliency}, depth cues~\cite{lang2012depth} etc. for estimating saliency. Also, Erdem \textit{et al.}~\cite{erdem2013visual} have proposed a non-linear feature integration approach for saliency prediction, using regional covariance features~\cite{porikli2006covariance}.

While the works discussed above are mainly driven by results from neuroscience and psychology, there have also been works which are motivated from an information theory perspective~\cite{kountchev2012advances}. Oliva \textit{et al.}~\cite{oliva2003top} have taken a top-down approach wherein the  statistical rarity of local features across the scene becomes a crucial factor for a region to be salient. Bruce \textit{et al.}~\cite{bruce2005saliency} have explored an information theoretic approach, where the self-information of local image content is used in predicting attention allocation.

Recent progress in saliency prediction has mostly been driven by the advances in deep learning. Convolutional Neural Networks (CNN), whose design is motivated by the functioning of cells in visual cortex of primates, can capture semantically rich visual features in a hierarchical fashion. While some of these works extract features from pre-trained CNNs~\cite{kummerer2014deep}, a few others have trained their networks specifically for saliency prediction~\cite{vig2014large}~\cite{liu2015predicting}.

In contrast to the traditional usage of multi-scale hand crafted image features, K\"ummerer \textit{et al.}~\cite{kummerer2014deep} have proposed an approach of using feature representations from AlexNet~\cite{krizhevsky2012imagenet} trained for object classification, to perform saliency prediction. Further, the extensive experimental evaluation conducted by K\"ummerer \textit{et al.} highlights the importance of feature representations from deeper layers of a CNN in saliency prediction. Though CNN representations are usually generalizable between vision tasks (e.g., object classification to saliency prediction)~\cite{razavian2014cnn}, our work emphasises that a task-specific convolutional neural network with a large depth, trained in an end-to-end manner, can outperform approaches using off-the-shelf CNN features by a large margin.

A recent work by Vig \textit{et al.}~\cite{vig2014large} proposes a method for obtaining optimal features by performing a large scale search over different feature-generating model configurations. Each model is considered to be a small convolutional neural network with a maximum of 3 layers, to constrain the computational complexity of overall system. However, this model can not efficiently leverage the semantic feature extracting capabilities of CNNs because of the small depth $(<=3)$ of the individual feature extractors. 

Liu \textit{et al.}~\cite{liu2015predicting} construct an ensemble of CNNs, termed as Multiresolution-CNN, for predicting eye fixations. Each of these CNNs is trained to classify image patches, at a particular scale, for saliency. This approach of predicting saliency with multiple scale-specific CNNs is efficient for capturing both the low-level and high-level aspects of an image. However, since this approach presents isolated image patches to the network, it fails to capture the global context, which is often crucial for saliency. The proposed DeepFix network overcomes this by operating on the image as a whole and capturing the semantics at multiple scales through its inception style convolution blocks.

\section{Proposed Approach}

\subsection{DeepFix Architecture}

\begin{figure}
\centering
\includegraphics[width=\linewidth]{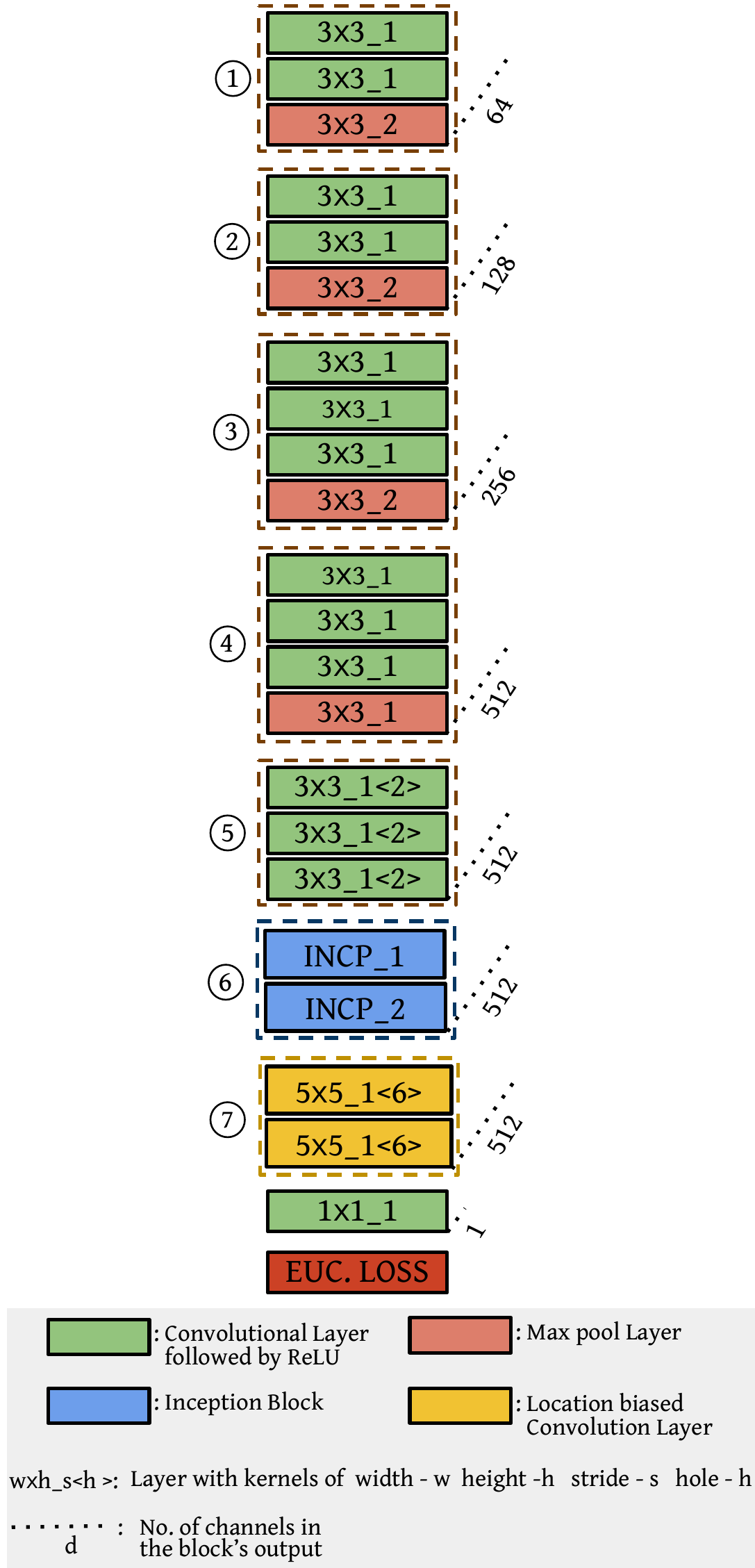}
\caption{Architecture of the proposed DeepFix model}
\label{fig:DeepFixArchitecture}
\end{figure}

DeepFix is a fully convolutional neural network, trained to predict pixel-wise saliency values for a given image in an end-to-end manner. The key features of the proposed CNN architecture are described below:\\

\begin{enumerate}
 
  \item The network takes an image of size $W\times H\times 3$ (RGB image) as input. This is followed by a series of 5 convolution blocks (\circlednumber{1} to \circlednumber{5}), depicted as dashed brown boxes in Fig.~\ref{fig:DeepFixArchitecture}.
  
  \item Similar to the VGG-16 net~\cite{simonyan2014very}, the first two blocks (\circlednumber{1}, \circlednumber{2}) contain two convolutional layers each, while the next three (\circlednumber{3}, \circlednumber{4}, \circlednumber{5}) have three convolutional layers. Each convolution filter in these five blocks is restricted to a kernel size of $3\times 3$ and operates with a stride of $1$. This allows the network to have a large depth without increasing the memory requirement. All convolutional layers in the network are followed by Rectified Linear Unit (ReLU) activation to introduce element wise non-linearity. 
  
  \item Additionally, each of the first four blocks (\circlednumber{1} to \circlednumber{4}) have a max-pool layer (of size $3\times3$) following the convolutional layers. Apart from introducing local translational invariance in its output, max-pooling (with stride) reduces computation for deeper layers while preserving the important activations~\cite{DeepLearning.net}. 
  
  Starting from the first convolutional block (\circlednumber{1}), the number of channels in the outputs of successive blocks gradually increase as $64, 128, 256,512$, depicted as numbers over dotted lines in Fig.~\ref{fig:DeepFixArchitecture}. This enables the net to progressively learn richer semantic representations of the input image. However, to limit the overall blob size, the spatial dimensions of the blob are halved with every increase in the blob's depth. This is achieved by introducing a stride of $2$ in the max-pool layers of the first $3$ blocks. This results in an output blob of spatial dimensions $\frac{W}{8}\times \frac{H}{8}$ after the third block. These spatial dimensions are retained in all the further layers.
  
\begin{figure}
\centering
\includegraphics[width=\linewidth]{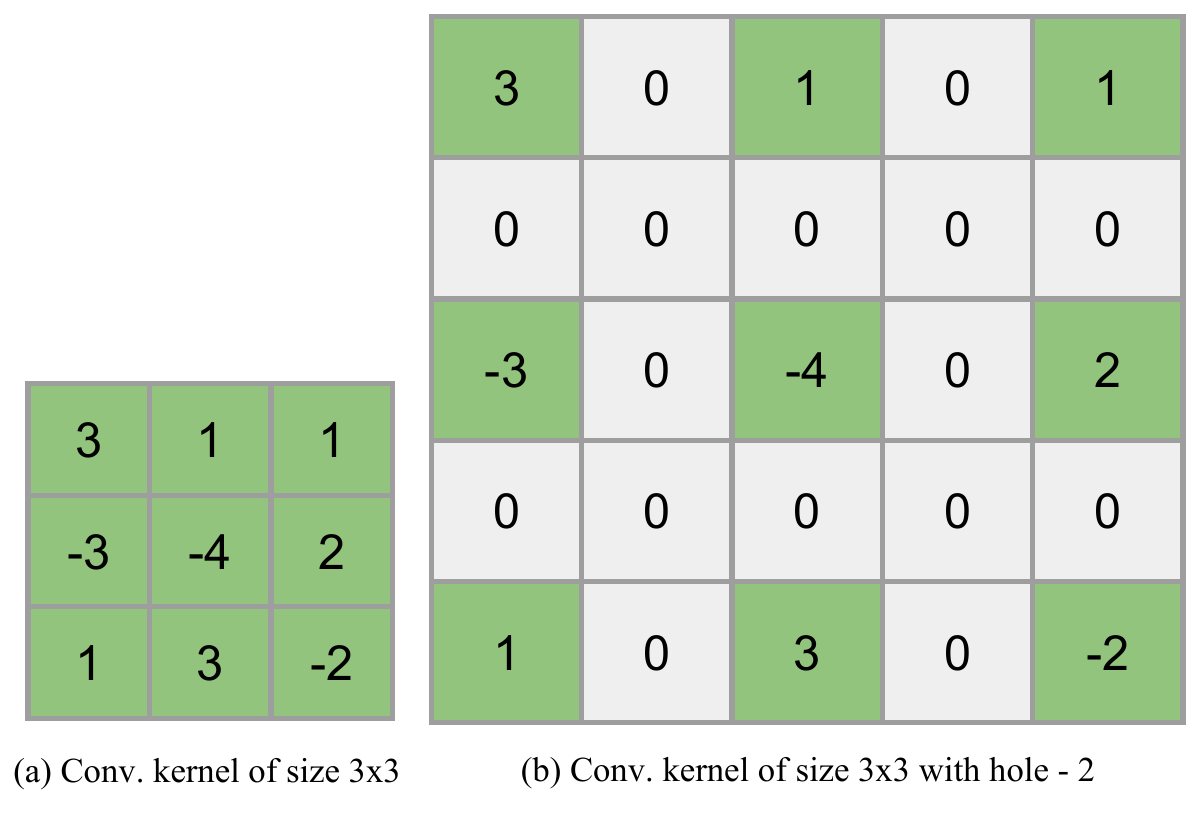}
\vspace{0.5cm}
\caption{Convolution kernel with holes. These kernels enables the layer to have a greater receptive field without increasing the memory footprint.}
\label{fig:holeFilter}
\end{figure}
  
  \item While training, the weights of the filters in the five convolution blocks (\circlednumber{1} to \circlednumber{5}) of DeepFix are initialized from the VGG-16 net. The weights of VGG-16 net have been learnt by training on 1.3 million images of the ImageNet~\cite{ILSVRC15} database. Initializing the weights from network trained on such a large corpus of images is observed to be important for stable and effective learning. 
  
  However, in the VGG-16 net, the spatial dimensions of the output blob are halved at the end of each convolution block, including the fourth and fifth blocks. In our network, to allow for the filters in the fifth block (\circlednumber{5}) to operate on the same receptive field they are originally trained for, we introduce holes of size 2 in their kernels. Convolution filters with holes are illustrated in Fig. \ref{fig:holeFilter}. The convolution filters of the fifth block, with kernel size $3\times3$ and hole size $2$ have a receptive field of $5\times5$. Jay \textit{et al.}~\cite{chen2014semantic} follow a similar procedure of introducing holes in filters to facilitate weight initialization in their work of semantic image segmentation.

  \item While the initial convolution blocks extract low-level image cues such as colour, contrast, texture, etc., the feature maps obtained from the fifth block are shown to characterize high-level semantic information~\cite{chatfield2014return}. Previous works have shown that saliency is best captured when semantics are considered from multiple scales~\cite{ittiModel, li2015visual}. Inspired by the recent success of GoogLeNet~\cite{szegedy2014going}, we capture the multi-scale semantic structure using two inception style convolution modules (in \circlednumber{6}), illustrated in Fig.~\ref{fig:Inception}.
  
  Each inception module operates on its input feature map with filters of multiple sizes, thereby capturing the multi-scale information. In the proposed inception module, we use convolutional layers of two sizes : $1\times1,\: 3\times 3.$ Unlike the inception module of GoogLeNet, the $5\times 5$ convolutional layer is simulated through a $\: 3\times 3$ layer with holes of size $2$. The number of channels in previous layer's output are brought down using $1\times1$ layers before feeding it to these layers. Additionally, a parallel max-pool layer is added in this inception module to bring in local invariance and it is followed by a $1\times1$ convolutional layer.
  
\begin{figure}
\centering
\includegraphics[width=\linewidth]{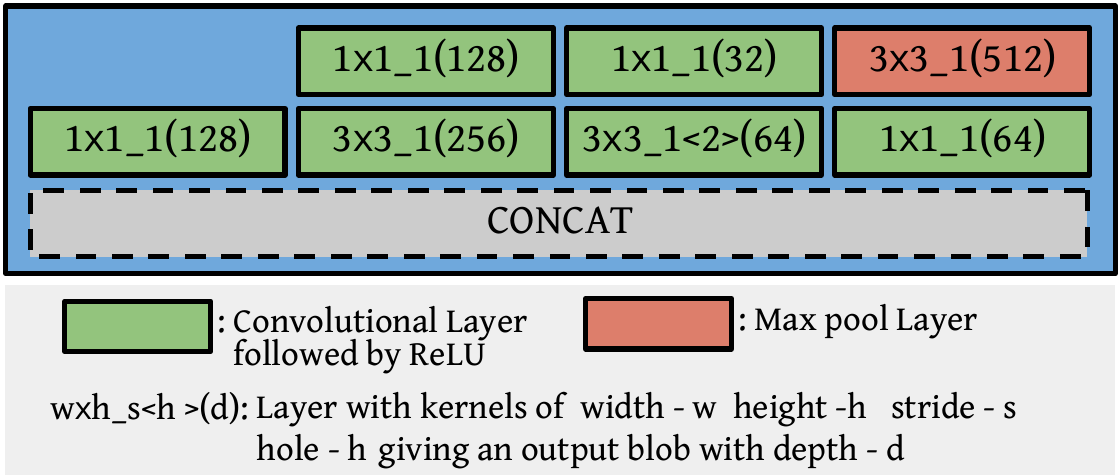}
\caption{Constituent layers of the Inception module used in DeepFix network}
\label{fig:Inception}
\end{figure}
  
  \item Saliency, in neuroscience literature, is often characterized as the unique quality of an entity by which it stands out with respect to its neighbours~\cite{Itti:2007}. This property can be best captured when the local semantic features of an image region are examined in the context of its neighborhood. To facilitate this, the convolutional layers (in \circlednumber{7}) following the inception modules are designed to have very large receptive fields by introducing holes of size $6$ in their convolution kernels. As shown in Fig.~\ref{fig:holeFilter}, these filters with holes can operate on input regions larger than their actual kernel size without increasing the memory footprint. These two layers have a receptive field of $25 \times 25$ on their respective inputs.
  
  In addition, the convolution operation in these two layers is made location dependent to model the centre-bias observed in eye fixations. We term these layers as \textit{Location Biased Convolutional} (LBC) layers and are explained in detail in Sec.~\ref{subsec:LBC}. We observe that these two layers are effective at capturing the global context of the image and have resulted in a significant performance improvement.
  
  To avoid over-fitting and to make the model more general, we have introduced drop-out in the output of the second LBC layer. We have chosen a dropout ratio of $0.5$.
  
  \item Finally, the output from the second LBC layer (in (\circlednumber{7}) is fed to a $1\times1$ convolutional layer whose output is taken to be the predicted saliency map.  The obtained map has a spatial resolution of $\frac{W}{8}\times\frac{H}{8}$ due to the greater-than-unit stride present in the max-pool layers of the first three blocks. We upsample this map to the original image resolution using bicubic interpolation.
  
\end{enumerate}

\subsection{Centre-bias in Eye Fixations}
\label{subsec:LBC}

Statistically, it has been observed that a significant number of the human eye fixations fall in the central region of an image. This tendency of humans to gaze at the centre while free-viewing is termed as centre-bias in eye fixations and has been studied extensively in the fields of neuroscience and psychology~\cite{tatler2007central, tseng2009quantifying}. This phenomenon is often explained through photographer's bias - the innate tendency of photographers to capture the object of interest by positioning it at the centre of the view. This is found to result in a secondary effect where viewers, after repeatedly viewing such images with the photographer's bias, expect that the most informative content of an image is likely to be present at the centre~\cite{tseng2009quantifying}. This guides their attention involuntarily towards central region of an image even in the absence of the photographer's bias. This secondary effect is described in the literature as bias due to viewing strategy. Also, Borji \textit{
et al.}, in their work on CAT2000~\cite{CAT2000}, experimentally observed that uninterestingness of images could result in eye fixations towards the image centre. These findings suggest that eye fixation patterns are an outcome of both the underlying stimulus and its location. 

To account for the spatial biases present in the human eye fixations, some works in the past have employed an explicit centre-bias term in their saliency prediction models~\cite{vig2014large, kummerer2014deep, yang2010chance}. In contrast, we design the DeepFix architecture to \textit{learn} location-dependent patterns implicitly through a novel Location Biased Convolutional (LBC) layer thereby accounting for biases effectively in an image-dependent manner. Next, we describe the construction of LBC layers.

\begin{figure}
\centering
\includegraphics[width=\linewidth]{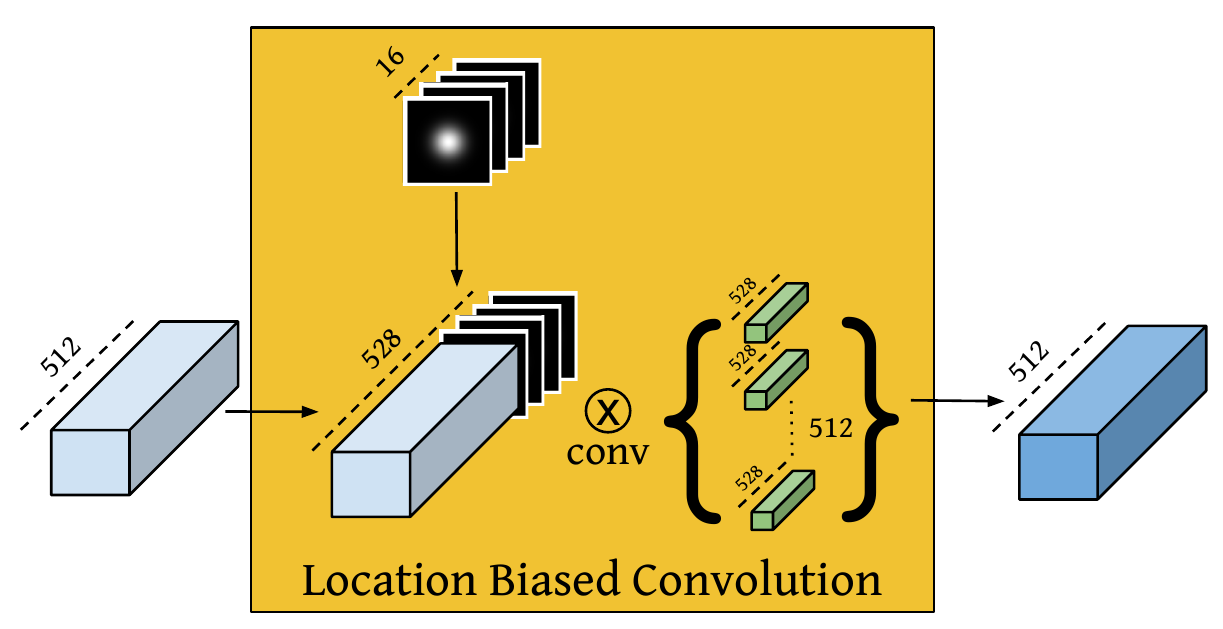}
\caption{Location Biased Convolution Filter for learning location dependent patterns in data (e.g., centre-bias present in the eye-fixations). The usual bias term associated with convolutional layers and the ReLU activation are assumed to be present and are not shown explicitly in the above diagram.}
\label{fig:LBC}
\end{figure}

\subsubsection{Location Biased Convolutional (LBC) layer}
\label{subsubsec:LBC}

The constituent layers of a fully convolutional net (Convolutional, ReLU and Max-pool layers) are location invariant i.e., a spatial shift of input image will only result in an equivalent scaled spatial shift in the output while keeping the values nearly intact. This property of fully convolutional nets prevents them from learning any location-dependent patterns such as the centre-bias. We tackle this problem by introducing location dependency in the two convolutional layers following the inception modules of the proposed DeepFix architecture. 

Let $\boldsymbol{W_c}$ represent weights from $c^{th}$ filter in a convolutional layer and $b_c$ represent its bias. Let the feature vector at spatial location $(x, y)$ in the input blob to this layer be $\textbf{I}(x,y)$ and the $c^{th}$ filter's response be $R_c(x,y)$. This convolution operation can be represented as 

\vspace{-0.25cm}
\begin{equation}
R_c(x,y) = \mathcal{R}\Bigg(\sum\limits_{i,j} \Big(\textbf{I}(x+i,y+j)*\boldsymbol{W_c}(i,j) + b_c\Big)\Bigg)
\end{equation}

where $*$ represents dot product and $\mathcal{R}$ represents ReLU non-linearity.

Here, the weights $\boldsymbol{W_c}$ and the bias $b_c$ are completely independent of the location $(x,y)$ at which they operate, making the convolutional operation location invariant. Introducing spatial dependency directly by making the filter weights a function of the spatial coordinates will increase the number of layer parameters dramatically (proportional to the product of the spatial dimensions of input blob). This also goes against the principle of weight sharing in CNNs which is considered to be an important reason for their effectiveness in visual recognition. We address this problem by concatenating a data independent and location specific feature $\textbf{L}(x,y)$ to the existing input feature $\textbf{I}(x,y)$. This results only in a minimal increase in the number of layer parameters (i.e., linear with the dimension of the $\textbf{L}(x,y)$) and is independent of the input blob's spatial size. 

\vspace{-0.25cm}
\begin{equation}
\begin{split}
R_c(x,y) = \mathcal{R}\Bigg(\sum\limits_{i,j} \Big(\textbf{I}(x+i,y+j)*\boldsymbol{W_c}(i,j) + \\ \textbf{L}(x+i,y+j)*\boldsymbol{W^{'}}_c(i,j) + b_c\Big)\Bigg)
\end{split}
\end{equation}

\begin{figure}
\centering
\includegraphics[width=0.85\linewidth]{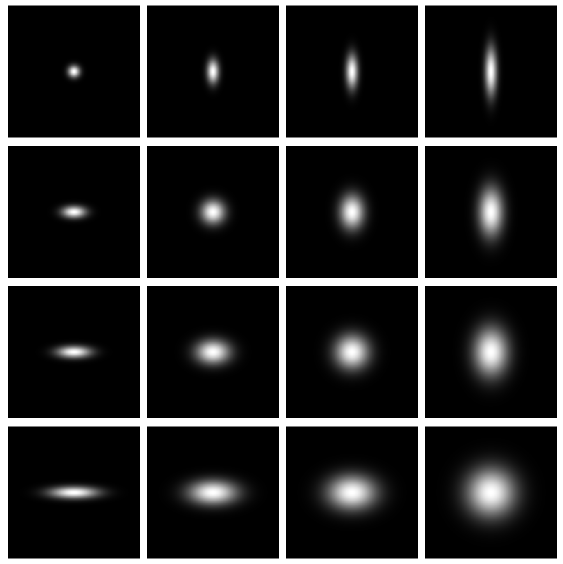}
\caption{Gaussian blobs with different horizontal and vertical variances concatenated to the input blob of LBC layers to make the layer's response location specific.}
\label{fig:centreBlobs}
\end{figure}

While the location specific features, $\textbf{L}(x,y)$, remain constant through the entire training procedure, the weights of a filter operating on it, $\boldsymbol{W^{'}}_c$, are learnt over training. This enables the network to optimally combine input stimuli with its location information for predicting the final saliency map. We choose the location-specific feature $\textbf{L}(x,y)$ to be $16$ dimensional with each component giving different weights to the central region, each obtained from a Gaussian of a specific horizontal and vertical variance. The $16$ constant maps from which these location specific features are obtained are shown in Fig. \ref{fig:centreBlobs}

\section{Experimental Evaluation}

\subsection{Training DeepFix}
\label{subsec:train}

The parameters of the first five convolution blocks are initialized from the VGG-16 net~\cite{simonyan2014very}. The weights in the convolutional layers of the inception modules and the two LBC layers following them are drawn from a Gaussian distribution with zero mean and standard deviation of $0.01$, and the biases are set to $0$. The weights of the last convolutional layer are also drawn from a zero mean Gaussian distribution but with a standard deviation of $10$ and the bias is set to $0$.\par

During training, all the layers in the five convolution blocks, whose weights are initialized from VGG-16, are learnt with an initial learning rate of $2 \times 10^{-4}$. The remaining layers, whose weights are randomly initialized from Gaussian, are assigned a higher learning rate of $2 \times 10^{-3}$. We scale down the learning rates of all the layers by a factor of $5$ whenever the performance saturates on the validation set. The network parameters are learned by back-propagating the euclidean loss of the predicted saliency map with respect to the ground truth saliency using Stochastic Gradient Descent (SGD) with momentum. The network is trained with a momentum of $0.9$ and a weight decay of $0.0005$.

We train our network in two stages. In the first stage, we use a mouse contingency based saliency dataset - SALICON~\cite{jiang2015salicon} for training. Though these saliency maps do not correspond to actual eye fixations, this dataset contains 15000 images, from various indoor and outdoor scenarios, providing a rich description of the problem to the CNN. In the second stage, we train the network using smaller datasets, with ground-truth saliency maps generated from actual eye fixation data. We evaluate our network by testing on two datasets - CAT2000~\cite{CAT2000}, MIT300~\cite{Judd_2012}. The network used for testing on CAT2000 is trained in the second stage with CAT2000 train set while the network used for testing on MIT300 is trained with images from MIT1003~\cite{judd2009learning} dataset. The entire training procedure takes about $1$ day on a K40 GPU with the caffe deep learning framework~\cite{jia2014caffe}.\par

Now we will briefly describe the 4 saliency datasets used during the training and testing phases of the DeepFix network.

\vspace{0.2cm}
\textbf{SALICON : }SALIency in CONtext (SALICON) dataset~\cite{jiang2015salicon} contains $10,000$ training images, $5,000$ validation images and $5,000$ test images for saliency prediction. The authors of SALICON~\cite{jiang2015salicon} propose mouse-contingent-tracking on multi-resolution images as an effective replacement for eye-contingent-tracking in saliency map generation. Further, they show, both qualitatively and quantitatively, that a high degree of similarity exists between the two. With their method, a large mouse-tracking based saliency dataset of 20,000 images has been created from MS COCO~\cite{lin2014microsoft}. By far, this is the largest selective attention dataset with images from varying context~\cite{jiang2015salicon}. In our work, we have used $15,000$ images ($10,000$ training images$+5,000$ validation images) during the first stage training of the DeepFix.

\vspace{0.2cm}
\textbf{CAT2000 :} This dataset~\cite{CAT2000} contains 4000 images selected from a wide variety of image categories such as \textit{Cartoons, Art, Satellite, Low resolution images, Indoor, Outdoor, Jumbled, Random, and Line drawings} etc.. Overall, this dataset contains 20 different categories with 200 images from each category. The saliency maps for 2000 images (100 from each category) are released as a part of the train set while the saliency maps for the rest of the 2000 images are held-out and they form the test set. From the set of 2000 train images, we have used 1800 images (90 images from each category) during the second stage training of the CNN while the remaining 200 images (10 images from each category) are used for validation. After the second stage training on 1800 images, the proposed method is evaluated on the CAT2000 test set using the MIT saliency benchmark~\cite{mit-saliency-benchmark}.

\vspace{0.2cm}
\textbf{MIT1003 :} MIT1003~\cite{judd2009learning} dataset is a collection of 1003 random images from Flickr and LabelMe whose saliency maps are generated using eye tracking data from fifteen users. We use 900 of these images for the second stage of training (for evaluating on MIT300) and the remaining 103 images for validation.

\vspace{0.2cm}
\textbf{MIT300 :} This dataset~\cite{Judd_2012} contains 300 natural images from both indoor and outdoor scenarios. The ground-truth for this entire dataset is held-out and we use this for evaluating the proposed DeepFix model using the MIT saliency benchmark~\cite{mit-saliency-benchmark}. 

\begin{figure*}
\centering
\vspace{0.5cm}
\subfloat{\hspace{0.8cm}Image\hspace{1.6cm}GBVS\cite{harel2006graph}\hspace{1.0cm} eDN\cite{vig2014large} \hspace{1.0cm} BMS\cite{zhang2013saliency} \hspace{0.6cm} Mr-CNN\cite{liu2015predicting} \hspace{0.05cm} \textbf{DeepFix(proposed)} \hspace{0.0cm} Ground-truth}\vspace{0.3cm}
\subfloat{1)  \includegraphics[width = 0.965\linewidth]{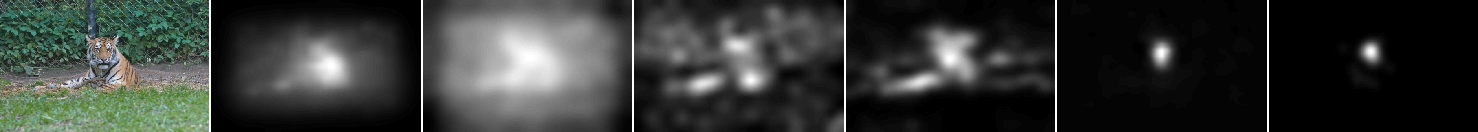}}\vspace{-0.2cm}
\subfloat{2)  \includegraphics[width = 0.965\linewidth]{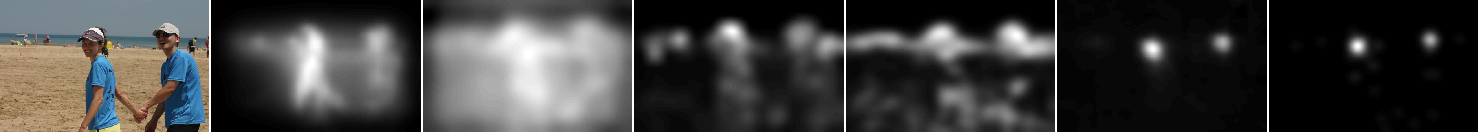}}\vspace{-0.2cm}
\subfloat{3)  \includegraphics[width = 0.965\linewidth]{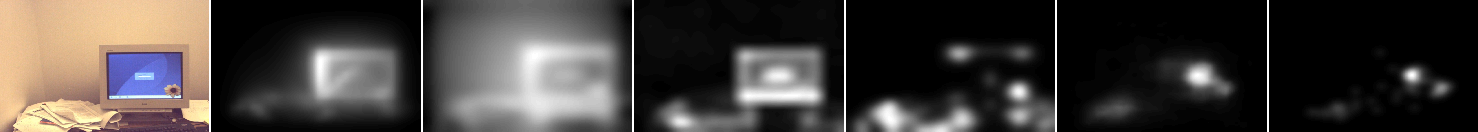}}\vspace{-0.2cm}
\subfloat{4)  \includegraphics[width = 0.965\linewidth]{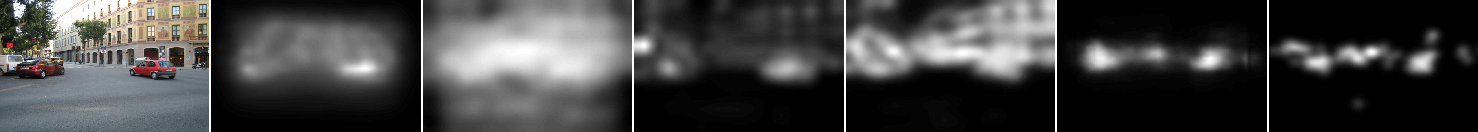}}\vspace{-0.2cm}
\subfloat{5)  \includegraphics[width = 0.965\linewidth]{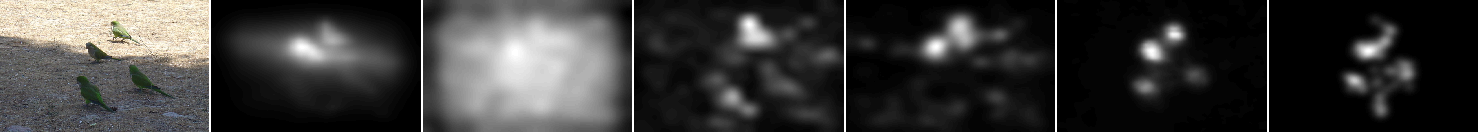}}\vspace{-0.2cm}
\subfloat{6)  \includegraphics[width = 0.965\linewidth]{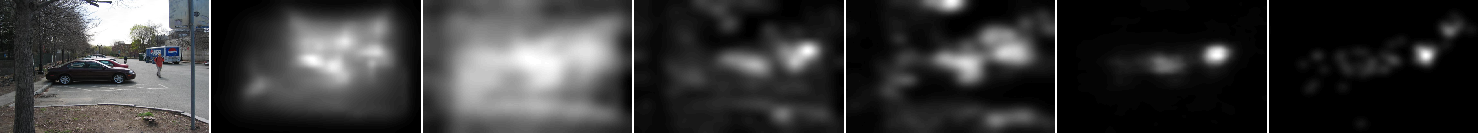}}\vspace{-0.2cm}
\subfloat{7)  \includegraphics[width = 0.965\linewidth]{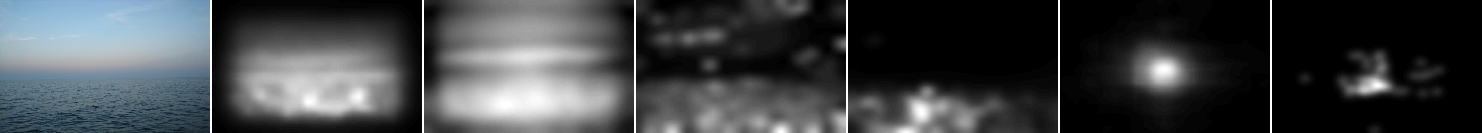}}\vspace{-0.2cm}
\subfloat{8)  \includegraphics[width = 0.965\linewidth]{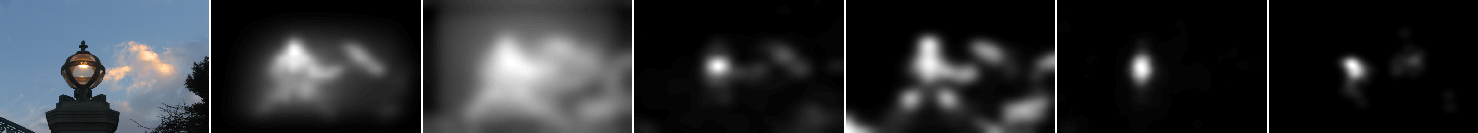}}\vspace{-0.2cm}
\subfloat{9)  \includegraphics[width = 0.965\linewidth]{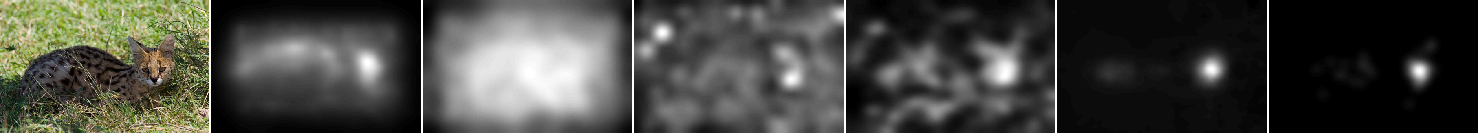}}\vspace{-0.2cm}
\subfloat{10) \includegraphics[width = 0.965\linewidth]{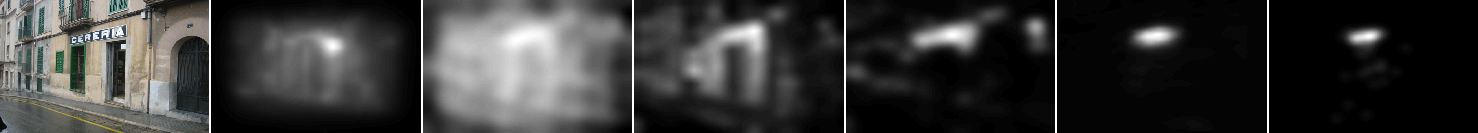}}\vspace{-0.2cm}
\subfloat{11) \includegraphics[width = 0.965\linewidth]{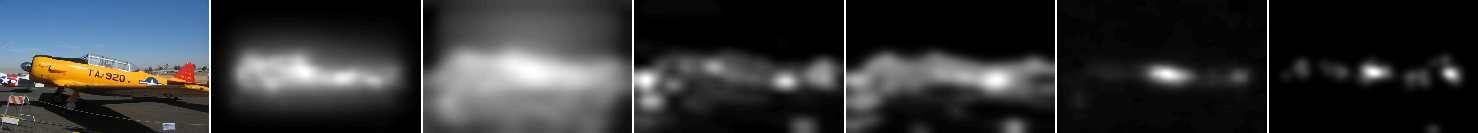}}\vspace{-0.2cm}
\subfloat{12) \includegraphics[width = 0.965\linewidth]{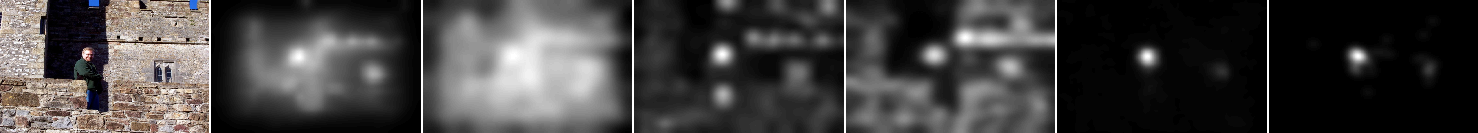}}\vspace{-0.2cm}
\vspace{0.5cm}
\caption{Qualitative results of the proposed method on validation images from MIT1003 dataset}
\label{fig:results_qualitative}
\end{figure*}

\subsection{Evaluation}

We have evaluated the performance of our network on the MIT Saliency Benchmark~\cite{mit-saliency-benchmark} with held-out test sets of MIT300 and CAT2000. The benchmark evaluates models on a variety of metrics, namely Earth Mover's Distance (EMD), Normalized Scanpath Saliency (NSS), Similarity, Linear Correlation Coefficient (CC), AUC-Judd, AUC-Borji, shuffled-AUC. Previous studies on saliency metrics by Riche ~\textit{et al.}~\cite{riche2013saliency} and Borji \textit{et al.}~\cite{peters2008applying} show that evaluating saliency models by any one of these metrics does not result in a fair comparison which can reflect the qualitative results. Here, we briefly describe these evaluation metrics:

Let $G$ denote the ground-truth saliency map of an image and $S$ be the map estimated using a saliency prediction model.
\vspace{0.2cm}

\textbf{Earth Mover's Distance (EMD) : } EMD is a measure of the distance between the two 2D maps, $G$ and $S$. It is the minimal cost of transforming the probability distribution of the estimated saliency map $S$ to that of the ground truth map $G$. Therefore, the lesser the EMD score, the better is the estimated saliency map.

\textbf{Normalized Scanpath Saliency (NSS) :} Normalized Scanpath Saliency is a metric specifically introduced for saliency map evaluaiton by Peters \textit{et al.}~\cite{peters2005components}. This metric is calculated by taking the mean of scores assigned by the unit normalized saliency map $S_\mathcal{N}$ (with zero mean and unit standard deviation) at human eye fixations.

\begin{equation}
    NSS = \frac{1}{N}\sum_{i=1}^{N}S_\mathcal{N}(i)
\end{equation}

Here, N denotes the number of human eye positions.

\textbf{Linear Correlation Coefficient(CC) :} The correlation coefficient metric between $G$ and $S$ is given by :

\begin{equation}
    CC = \frac{cov(G,S)}{\sigma_G*\sigma_S}
\end{equation}

It gives a measure of the linear relationship between the two maps. A score close to $+1$ indicates almost a perfect linear relationship between the maps.\par

\textbf{Similarity metric :} Similarity metric computes the sum of the minimum values at each pixel location for the two distributions($S_\mathscr{N}$ and $G_\mathscr{N}$). Here, $S_\mathscr{N}$ and $G_\mathscr{N}$ are normalized to be probability distributions.

\begin{equation}
   Sim = \sum_{i=1}^{N}min(S_\mathscr{N}(i),G_\mathscr{N}(i))
\end{equation}
\begin{equation}
  \sum_{i=1}^{N}S_\mathscr{N}(i)=1   \hspace{1cm} \sum_{i=1}^{N}G_\mathscr{N}(i)=1
\end{equation}

Here $S_\mathscr{N}$ and $G_\mathscr{N}$ are the normalized distributions and  $N$ denotes all the pixel locations in the 2D maps. As the name suggests, a score of 1 denotes that the two maps are the same.

\textbf{Area Under Curve (AUC) :} Area Under the ROC curve (AUC) is one of the widely used metrics for the evalaution of the maps estimated by saliency models. In AUC, two image locations are used : the actual human fixations as the positive set (fixation distribution) and some points randomly sampled from the image as the negative set (non-fixation distribution). Depending upon the choice of the non-fixation distribution, there are mainly two versions of AUC : AUC with uniform distribution of non-fixation points (AUC-Judd and AUC-Borji) and the shuffled-AUC. The shuffled-AUC uses human fixations of other images in the dataset (to take into account the center-bias) as the non-fixation distribution. Thus shuffled-AUC tends to give a lower score to those models which explicitly incorporate center-bias~\cite{barthelme2013modeling}.

\begin{table*}[!t]
\centering
\caption{Experimental Evaluation on CAT2000 Test Set}
\begin{tabular}{|l||l|l|l|l|l|l|l|l|}
\hline
Saliency Models &AUC-Judd $\uparrow$  &SIM $\uparrow$  &EMD $\downarrow$ &AUC-Borji $\uparrow$ &shuffled AUC $\uparrow$ &CC $\uparrow$ &NSS $\uparrow$ \\ \hline\hline
DeepFix (Proposed)  &\textbf{0.87} &\textbf{0.75} &\textbf{1.11} &0.81 &0.57 &\textbf{0.88} &\textbf{2.29}\\ \hline
Context Aware Saliency \cite{cheng2015global}&0.77 &0.50 &3.09 &0.76 &\textbf{0.60} &0.42 &1.07 \\ \hline
Judd Model \cite{judd2009learning}&0.84 &0.46 &3.61 &\textbf{0.84} &0.56 &0.54 &1.30 \\ \hline
GBVS \cite{harel2006graph}&0.80 &0.51 &2.99 &0.79 &0.58 &0.50 &1.23\\ \hline
\end{tabular}
\vspace{5mm}
\label{tab:results_cat2000}
\end{table*}

\begin{table*}[!t]
\centering
\caption{Experimental Evaluation on MIT300 Test Set}
\begin{tabular}{|l||l|l|l|l|l|l|l|l|}
\hline
Saliency Models &AUC-Judd $\uparrow$  &SIM $\uparrow$  &EMD $\downarrow$ &AUC-Borji $\uparrow$ &shuffled AUC $\uparrow$ &CC $\uparrow$ &NSS $\uparrow$ \\ \hline\hline
DeepFix (Proposed)  &\textbf{0.87} &\textbf{0.67} &\textbf{2.04} &0.80 &0.71 &\textbf{0.78} &\textbf{2.26}\\ \hline
SALICON \cite{salicon}&\textbf{0.87} &0.60 &2.62 &\textbf{0.85} &\textbf{0.74} &0.74 & 2.12 \\ \hline
Mr-CNN \cite{liu2015predicting} &0.77 &0.45 &4.33 &0.76 &0.69 &0.41 &1.13 \\ \hline
Deep Gaze I \cite{kummerer2014deep}&0.84 &0.39 &4.97 &0.83 &0.66 &0.48 &1.22 \\ \hline
BMS\cite{zhang2013saliency}  &0.83 &0.51 &3.35 &0.82 &0.65 &0.55 &1.41 \\ \hline
eDN \cite{vig2014large}&0.82 &0.41 &4.56 &0.81 &0.62 &0.45 &1.14\\ \hline
Context Aware Saliency \cite{cheng2015global}&0.74 &0.43 &4.46 &0.73 &0.65 &0.36 &0.95 \\ \hline
Judd Model \cite{judd2009learning}&0.81 &0.42 &4.45 &0.80 &0.60 &0.47 &1.18 \\ \hline
GBVS \cite{harel2006graph}&0.81 &0.48 &3.51 &0.80 &0.63 &0.48 &1.24\\ \hline
\end{tabular}
\vspace{5mm}
\label{tab:results_mit300}
\end{table*}

\subsection{Results}

The qualitative results obtained by the DeepFix network, along with that of other recent methods on a few example images from MIT1003\footnote{As the ground-truth saliency maps of the test sets are not publicly available, the qualitative results are shown on the MIT1003 validation set.} are shown in Fig. \ref{fig:results_qualitative}. As shown in the figure, the proposed network is able to efficiently capture saliency arising from both low-level features such as colour variations (rows 3, 5), shape (row 8) etc. as well as the more high-level aspects such as text (rows 10, 11), faces of animals (rows 1, 9), and humans (row 12). In the case of images without any strikingly salient regions (row 7), the network relies on the learnt location pattern of humans gazing towards the centre to predict saliency accurately. The network is also able to analyze the relative importance of these factors and weigh them appropriately in the final saliency map. For instance, in row 2 of Fig. \ref{fig:results_qualitative}, 
DeepFix attributes saliency not to the bright colored T-shirts worn by the people (low-level), but to their faces (high-level). We attribute the performance of the proposed CNN architecture to its large depth allowing it to learn richer semantic representations, filters capturing multi-scale semantics and global context, as well as the implicit learning of location dependent patterns in eye fixations.

\begin{figure*}
\centering
\includegraphics[width = \linewidth]{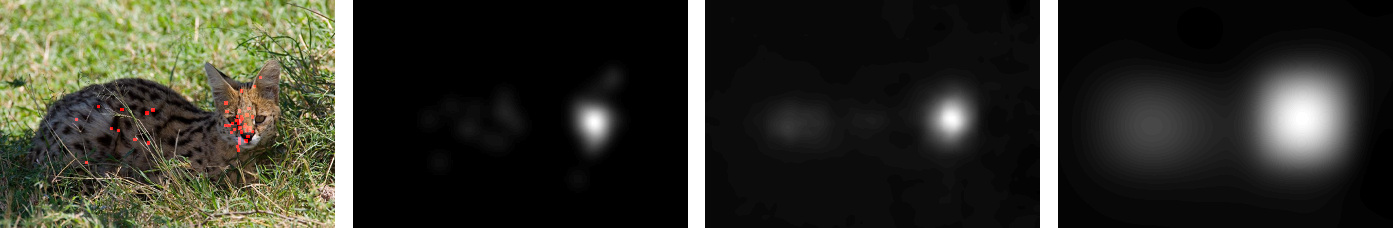}
\begin{center}
(a)\hspace{4.3cm}(b)\hspace{4.3cm}(c)\hspace{4.3cm}(d)
\end{center}
\caption{Illustration for the limitation of AUC metrics in penalizing false positives. For the image (a), with ground-truth saliency map (b), we have calculated the metric scores for two predictions (c) and (d). While (c) matches closely to the ground-truth, (d) can be observed to be highly blurred. The scores obtained for (c) are EMD = 1.04, NSS = 4.95, shuffled AUC = 0.88, AUC-Borji = 0.91. On the other hand, the scores obtained for (d) are EMD = 1.48, NSS = 3.23, shuffled AUC = 0.88, AUC-Borji = 0.94. It can be observed that the false positives in (d) are penalized by EMD and NSS metrics significantly. However, the shuffled AUC assigns the same score for both (c), (d) and contrary to expectations, AUC-Borji assigns a higher score for (d) than for (c). This limitation of AUC measures for saliency prediction has been observed before by ~\cite{borji2013analysis}\cite{zhao2011learning}.}
\label{fig:AUC}
\end{figure*}

 The quantitative results obtained on the test sets of CAT2000 and MIT300 are presented in Tables. \ref{tab:results_cat2000} and \ref{tab:results_mit300} respectively\footnote{The results of SALICON are taken from the MIT Saliency Leaderboard~\cite{mit-saliency-benchmark}. However, this model and paper have not been made publicly available at the time of writing.}. The results obtained show that the proposed method achieves state-of-the-art results on both the datasets. As evident from Tables. \ref{tab:results_cat2000} and  \ref{tab:results_mit300}, our approach outperforms other methods by a huge margin with respect to a majority of the metrics -- NSS, EMD, CC and Similarity, on both the datasets. The proposed method does not show a similar gain in performance on the AUC metrics. The AUC metrics reward methods primarily based on true positives, while the false positives generated do not incur heavy penalties. This can often result in blurred/hazy saliency maps receiving good scores as shown in Fig.~\ref{fig:AUC}. This drawback of AUC metrics has been previously observed by Borji \textit{et al.}~\cite{borji2013analysis} and Zhao \textit{et al.}~\cite{zhao2011learning}. 
 
 The AUC-Shuffled metric is specifically designed to penalize models which account for the centre-bias present in eye fixations. Since our network learns this centre-bias as a location dependent fixation pattern, we obtain lower scores on shuffled AUC, despite the qualitative similarity of the predicted saliency maps with the ground-truth. For example, for the image in row 7 of Fig.~\ref{fig:results_qualitative}, the saliency map predicted by DeepFix receives a shuffled AUC score of $0.68$ where as the map predicted by eDN~\cite{vig2014large} receives a higher score of $0.70$.

\begin{table*}[!t]
\centering
\caption{Ablation Analysis of the proposed method on MIT1003 validation set}
\begin{tabular}{|l||l|l|l|l|l|l||l|}
\hline
Method &AUC-Judd $\uparrow$  &SIM $\uparrow$  &EMD $\downarrow$ &AUC-Borji $\uparrow$ &shuffled AUC $\uparrow$ &CC $\uparrow$ &NSS $\uparrow$ \\ \hline\hline

\hline\hline

DF-No-LBC & 0.90 & 0.52 & 1.45 & 0.85& \textbf{0.75}& 0.70& 2.54\\\hline
DF-Explicit-CB & 0.90 & 0.50 & 1.59 & \textbf{0.87}& 0.73& 0.69& 2.44\\\hline
DF-LBC & 0.90 & \textbf{0.54} & \textbf{1.28} & 0.85 & 0.74 & \textbf{0.72}& \textbf{2.58}\\\hline

\end{tabular}
\vspace{5mm}
\label{tab:ablation}
\end{table*}

\subsection{Ablation Analysis}

In this subsection, we analyze the effect of LBC layers in the saliency map prediction. For this analysis, we construct three variations of our model.

\begin{enumerate}

\item DeepFix with no LBC layers (DF-No-LBC): The data-independent feature concatenated to the input of LBC layers, discussed in Sec.~\ref{subsubsec:LBC}, is aimed at introducing location dependence in the convolution operation. We remove this from the proposed DeepFix architecture converting the LBC layers to the usual convolutional layers.

\item DeepFix with explicit centre bias (DF-Explicit-CB): We create a mean saliency map, by averaging saliency maps of all training images, to obtain the centre-bias present in the dataset. This mean map is added to the output of DF-No-LBC to generate the final saliency map.

\item DeepFix with LBC layers (DF-LBC): This is the proposed architecture described in Sec. III and shown in Fig.~\ref{fig:DeepFixArchitecture}.

\end{enumerate}

The above discussed models are trained as described in Sec.~\ref{subsec:train} and are evaluated on the validation set of MIT1003. The quantitative results obtained for the above three scenarios are presented in Table.~\ref{tab:ablation}. The results emphasize that implicitly learning location dependent patterns through LBC layers results in better saliency prediction. We also found that, in the case of DF-Explicit-CB, varying the relative weights of the mean map and the predicted map does not result in increase of the performance on any of the metrics other than AUC-Borji.

\section{Conclusion}
In this work, we have proposed a first-of-its-kind fully convolutional neural net - DeepFix for predicting human eye fixations on images. The proposed deep net utilizes the potential of the `inception module' to extract complex semantic features at multiple scales and also exploits the ability of `filters with holes' to capture global context via large receptive fields. We introduce \textit{LBC - Location Biased Convolutional filters}, a novel technique which enables the deep network to learn location dependent patterns. We show the advantage of LBC over traditional techniques of explicit bias addition, by means of an ablation analysis. Lastly, we show that an effective combination of the above mentioned concepts is able to outperform other state-of-the-art methods by a considerable margin.

{
\bibliographystyle{IEEEtran}
\bibliography{DeepFix.bib}
}

\end{document}